%% file: paper.tex
\documentclass[letterpaper]{article}
\usepackage{aaai}
\nocopyright
\usepackage{times}
\usepackage{helvet}
\usepackage{courier}
\usepackage{amsmath}
\usepackage[boxed,vlined]{algorithm2e}
\usepackage{graphicx}

\input{macro}

\title{Heuristics in Conflict Resolution}
\author{Christian Drescher \and Martin Gebser \and Benjamin Kaufmann \and Torsten Schaub\\
  Universit\"at Potsdam,  
  Institut f\"ur Informatik,
  August-Bebel-Str.\ 89,
  D-14482 Potsdam, Germany
}

\begin{document}

\maketitle
\input{abstract}

\input{introduction}
\input{background}

\input{algorithms}
\input{cnflanalysis}
\input{heuristics}
\input{experiments}
\input{conclusions}


\end{document}

%% file: macro.tex
\newtheorem{theorem}{Theorem} 

\newcommand{\head}[1]{\ensuremath{\mathit{head}(#1)}} 
\newcommand{\atom}[1]{\ensuremath{\mathit{atom}(#1)}}
\newcommand{\body}[1]{\ensuremath{\mathit{body}(#1)}} 

\newcommand{\naf}[1]{\ensuremath{{\sim}{#1}}}

\newcommand{\EB}[2]{\ensuremath{\mathit{EB}_{#2}(#1)}}

\restylealgo{algoruled}
\dontprintsemicolon
\linesnumbered
\SetVline
\SetFuncSty{textsc}

\SetKwInOut{Global}{Global}
\SetKwInOut{Input}{Input}
\SetKwInOut{Output}{Output}
\SetKw{Backtrack}{return}
\SetKw{Print}{print}
\SetKw{Return}{return}
\SetKw{Exit}{exit}
\SetKw{Or}{or}
\SetKwFunction{Select}{Select}
\SetKwFunction{SelectAntecedent}{SelectAntecedent}
\SetKwFunction{Propagation}{Propagation}
\SetKwFunction{UnitPropagation}{LocalPropagation}
\SetKwFunction{ConflictAnalysis}{ConflictAnalysis}
\SetKwFunction{Analysis}{ConflictAnalysis}
\SetKwFunction{NonTight}{NonTight}
\SetKwFunction{UnFoundedSet}{UnfoundedSet}
\SetKwFunction{LoopFormula}{LoopNogood}
\SetKwFor{Let}{let}{in}{tel}
\SetKwFor{Shortl}{let}{}{tel}
\SetKwFor{Such}{let}{such that}{tel}
\SetKwFor{Loop}{loop}{}{}
\SetCommentSty{textit}

\newcommand{\Tsign}{\ensuremath{\mathbf{T}}}
\newcommand{\Fsign}{\ensuremath{\mathbf{F}}}
\newcommand{\Tlit}[1]{\ensuremath{\Tsign #1}}
\newcommand{\Flit}[1]{\ensuremath{\Fsign #1}}

\newcommand{\Tass}[1]{\ensuremath{#1^{\Tsign}}}
\newcommand{\Fass}[1]{\ensuremath{#1^{\Fsign}}}

\newcommand{\sigmaUIP}{\sigma}
\newcommand{\sigmaAUX}{\rho}

\newcommand{\sysfont}[1]{\textit{#1}}
\newcommand{\chaff}[0]{\sysfont{chaff}}
\newcommand{\clasp}[0]{\sysfont{clasp}}
\newcommand{\cmodels}[0]{\sysfont{cmodels}}
\newcommand{\smodelscc}[0]{\sysfont{smodels}$_{\text{\sysfont{cc}}}$}

\newcommand{\claspsmall}[0]{\clasp$_{\mathit{short}}$}
\newcommand{\clasplex}[0]{\clasp$_{\mathit{lex}}$}
\newcommand{\claspavg}[0]{\clasp$_{\mathit{avg}}$}
\newcommand{\clasplcd}[0]{\clasp$_{\mathit{res}}$}
\newcommand{\claspactive}[0]{\clasp$_{\mathit{active}}$}
\newcommand{\claspprop}[0]{\clasp$_{\mathit{prop}}$}


%% file: abstract.tex
\begin{abstract}
Modern solvers for Boolean Satisfiability (SAT) and Answer Set Programming (ASP)
are based on sophisticated Boolean constraint solving techniques.
In both areas, conflict-driven learning and related techniques
constitute key features whose application is enabled by conflict analysis.
Although various conflict analysis schemes have been proposed, implemented,
and studied both theoretically and practically in the SAT area,
the heuristic aspects involved in conflict analysis have not yet received much attention.
Assuming a fixed conflict analysis scheme,
we address the open question of how to identify ``good'' reasons for conflicts,
and we investigate several heuristics for conflict analysis in ASP solving.
To our knowledge, a systematic study like ours has not yet been performed in the SAT area, thus,
it might be beneficial for both the field of ASP as well as the one of SAT solving.
\end{abstract}


%% file: introduction.tex
\section{Introduction}

The popularity of Answer Set Programming (ASP; \cite{baral02a}) as a paradigm for
knowledge representation and reasoning is mainly due to two factors:
first, its rich modeling language and,
second, the availability of high-performance ASP systems.
In fact, modern ASP solvers,
such as \clasp\ \cite{gekanesc07b}, \cmodels\ \cite{gilima06a}, and \smodelscc\ \cite{warsch04a},
have meanwhile closed the gap to Boolean Satisfiability
(SAT; \cite{mitchell05a}) solvers.
In both fields, conflict-driven learning and related techniques have led to
significant performance boosts \cite{baysch97a,marsak99a,momazhzhma01a,contest07a}.
The basic prerequisite for the application of such techniques is \emph{conflict analysis},
that is, the extraction of non-trivial reasons for dead ends encountered during search.
Even though ASP and SAT solvers exploit different inference patterns,
their underlying search techniques are closely related to each other.
For instance, the basic search strategy of SAT solver \chaff\ \cite{momazhzhma01a},
nowadays a quasi standard in SAT solving, is also exploited by ASP solver \clasp,
in particular, the principles of conflict analysis are similar.
Vice versa, the solution enumeration approach implemented in \clasp\ \cite{gekanesc07c}
could also be applied by SAT solvers.
Given these similarities, general search or, more specifically, conflict analysis
techniques developed in one community can (almost) immediately be exploited in
the other field too.
%

In this paper,
we address the problem of identifying ``good'' reasons for conflicts to be recorded
within an ASP solver.
In fact, conflict-driven learning exhibits several degrees of freedom.
For instance, 
several constraints may become violated simultaneously,
in which case one can choose the conflict(s) to be analyzed.
Furthermore, distinct schemes may be used for conflict analysis,
such as the resolution-based First-UIP and Last-UIP scheme \cite{zamamoma01a}.
Finally, if conflict analysis is based on resolution,
several constraints may be suitable resolvents, likewise permitting to eliminate some
literal in a resolution step.

For the feasibility of our study, 
it was necessary to prune dimensions of freedom in favor of predominant options.
In the SAT area, the \emph{First-UIP scheme} \cite{marsak99a} has empirically been shown to
yield better performance than other known conflict resolution strategies \cite{zamamoma01a}.
We thus fix the conflict analysis strategy to conflict resolution according to the First-UIP scheme.
Furthermore, it seems reasonable to analyze the first conflict detected by 
a solver (although conflicts encountered later on may actually yield ``better'' reasons).
This leaves to us the choice of the resolvents to be used for conflict resolution,
and we investigate this issue with respect to different goals:
reducing the size of reasons to be recorded,
skipping greater portions of the search space by backjumping (explained below),
reducing the number of conflict resolution steps, and 
reducing the overall number of encountered conflicts (roughly corresponding to runtime).
To this end, we modified the conflict analysis procedure of our ASP solver
\clasp\footnote{http://www.cs.uni-potsdam.de/clasp} for accommodating a
variety of heuristics for choosing resolvents.
The developed heuristics and comprehensive empirical results for them
are presented in this paper.


%% file: background.tex
\section{Logical Background}

We assume basic familiarity with answer set semantics (see, for instance, \cite{baral02a}).
This section briefly introduces notations and recalls a constraint-based
characterization of answer set semantics according to \cite{gekanesc07a}.
We consider propositional (normal) logic programs over an alphabet $\mathcal{P}$.
A \emph{logic program} is a finite set of \emph{rules} 
\begin{equation}\label{eq:rule}
 p_0 \leftarrow p_1, \dots, p_m, \naf{p_{m+1}}, \dots, \naf{p_n}
\end{equation}
where $0 \leq m \leq n$ and $p_i \in \mathcal{P}$ is an \emph{atom} for $0 \leq i \leq n$.
%
%
%
For a rule $r$ as in (\ref{eq:rule}), let $head(r) = p_0$ be the \emph{head}
of
$r$ and $body(r) = \{p_1, \dots, p_m, \naf{p_{m+1}}, \dots, \naf{p_n}\}$ be the
\emph{body} of~$r$.
The set of atoms occurring in a logic program $\Pi$ is denoted by $atom(\Pi)$,
and the set of bodies in $\Pi$ is $body(\Pi) = \{body(r) \mid r \in \Pi\}$.
For regrouping bodies sharing the same head $p$, define
$body(p) = \{body(r) \mid r \in \Pi, head(r) = p\}$.

For characterizing the answer sets of a program~$\Pi$,
we consider Boolean assignments $A$ over \emph{domain} $\mathit{dom}(A) =
atom(\Pi) \cup body(\Pi)$.
%
Formally, an  \emph{assignment} $A$ is a sequence
$(\sigma_1,\ldots,\sigma_n)$ of (signed) \emph{literals} $\sigma_i$ of the form
$\Tlit{v}$ or $\Flit{v}$ for $v\in \mathit{dom}(A)$ and $1 \leq i \leq n$.
Intuitively, $\Tlit{v}$ expresses that $v$ is \emph{true} and $\Flit{v}$ that it is \emph{false} in~$A$. 
We denote the complement of a literal $\sigma$ by
$\overline{\sigma}$, that is, $\overline{\Tlit{v}} = \Flit{v}$ and
$\overline{\Flit{v}} = \Tlit{v}$.
Furthermore, we let $A \circ B$ denote the
sequence obtained by concatenating two assignments $A$ and $B$.
We sometimes abuse notation and identify an assignment
with the set of its contained literals.
Given this, we access the true and false propositions in $A$ via
$\Tass{A} = \{p \in \mathit{dom}(A) \mid \Tlit{p} \in A\}$ and
$\Fass{A} = \{p \in \mathit{dom}(A) \mid \Flit{p} \in A\}$.
Finally, we denote the prefix of~$A$ up to a literal~$\sigma$ by
\begin{equation*}\label{eq:prefix}
A[\sigma]
=
\left\{
\begin{array}{@{}ll}
(\sigma_1,\dots,\sigma_m)
&
\text{if } 
A=(\sigma_1,\dots,\sigma_m,\sigma,\dots,\sigma_n) 
\\
A
&
\text{if }
\sigma\notin A
\ \text{.}
\end{array}
\right.
\end{equation*}

In our context, a \emph{nogood} \cite{dechter03} is a set $\{\sigma_1,\ldots,\sigma_m\}$ of literals,
expressing a constraint violated by any assignment containing $\sigma_1,\ldots,\sigma_m$.
An assignment $A$ such that $\Tass{A} \cup \Fass{A} = \mathit{dom}(A)$ and $\Tass{A} \cap
\Fass{A} = \emptyset$ is a \emph{solution} for a set~$\Delta$ of nogoods
if $\delta\not\subseteq A$ for all $\delta\in\Delta$.
Given a logic program~$\Pi$,
we below specify nogoods such that their solutions correspond to the answer sets of~$\Pi$.

We start by describing nogoods capturing the models of
the Clark's \emph{completion} \cite{clark78} of a program~$\Pi$.
For $(\beta=\{p_1, \dots, p_m, \naf{p_{m+1}},\dots,\naf{p_n}\})\in\body{\Pi}$, let
\begin{equation*}
\Delta_\beta
=
\left\{
\begin{array}{@{}l@{}}
\\[-3mm]
\{\Tlit{p_1},\dots,\Tlit{p_m},\Flit{p_{m+1}},\dots,\Flit{p_n},\Flit{\beta}\},
{} \\ 
\{\Flit{p_1},\Tlit{\beta}\},\dots,\{\Flit{p_m},\Tlit{\beta}\},
{} \\
\{\Tlit{p_{m+1}},\Tlit{\beta}\},\dots,\{\Tlit{p_n},\Tlit{\beta}\}
\\[0.5mm]
\end{array}
\right\}
\text{.}
\end{equation*}
Observe that every solution for $\Delta_\beta$ must assign body~$\beta$ 
equivalent to the conjunction of its elements.
Similarly,
for an atom $p\in\atom{\Pi}$,
the following nogoods stipulate $p$ to be equivalent to the disjunction
of $\body{p}=\{\beta_1,\dots,\beta_k\}$:
\begin{equation*}
\Delta_p
=
\left\{
\begin{array}{@{}l@{}}
\\[-3mm]
\{\Flit{\beta_1},\dots,\Flit{\beta_k},\Tlit{p}\},
{} \\ 
\{\Tlit{\beta_1},\Flit{p}\},\dots,\{\Tlit{\beta_k},\Flit{p}\}
\\[0.5mm]
\end{array}
\right\}
\text{.}
\end{equation*}
Combining the above nogoods for $\Pi$, we get
\begin{equation*}
\Delta_\Pi
=
\mbox{$\bigcup$}_{\beta\in\body{\Pi}}\Delta_\beta
\cup
\mbox{$\bigcup$}_{p\in\atom{\Pi}}\Delta_p
\text{ .}
\end{equation*}
The solutions for~$\Delta_\Pi$ correspond one-to-one to the
models of the completion of~$\Pi$.
If~$\Pi$ is \emph{tight}~\cite{fages94a,erdlif03a}, 
these models are guaranteed to match the answer sets of~$\Pi$.
This can be formally stated as follows.
%
\begin{theorem}[\cite{gekanesc07a}]\label{thm:nogoods:tight}
Let $\Pi$ be a tight logic program.
Then, $X\subseteq\atom{\Pi}$ is an answer set of~$\Pi$
iff
$X=\Tass{A}\cap\atom{\Pi}$ for a (unique) 
solution $A$ for $\Delta_\Pi$.
\end{theorem}

We proceed by considering non-tight programs~$\Pi$.
As shown in \cite{linzha04a}, \emph{loop formulas} can be added to the
completion of~$\Pi$ to establish full correspondence to the answer sets of~$\Pi$.
For $U\subseteq\atom{\Pi}$, let $\EB{U}{\Pi}$ be
\begin{equation*}
\{\body{r}\mid r\in\Pi, \head{r}\in U, \body{r}\cap U=\emptyset\}
\ \text{.}
\end{equation*}
Observe that $\EB{U}{\Pi}$ contains the bodies of all rules in~$\Pi$
that can \emph{externally support} \cite{lee05a} an atom in~$U$.
Given $U=\{p_1,\dots,p_j\}$ and $\EB{U}{\Pi}=\{\beta_1,\dots,\beta_k\}$, 
the following nogoods capture the loop formula of~$U$:
\begin{equation*}
\Lambda_U
=
\left\{
\begin{array}{@{}l@{}}
\\[-3mm]
\{\Flit{\beta_1},\dots,\Flit{\beta_k},\Tlit{p_1}\},\dots,
{} \\ 
\{\Flit{\beta_1},\dots,\Flit{\beta_k},\Tlit{p_j}\}
\\[0.5mm]
\end{array}
\right\}
\text{.}
\end{equation*}
Furthermore, we define
\begin{equation*}
\Lambda_\Pi
=
\mbox{$\bigcup$}_{U\subseteq\atom{\Pi}}\Lambda_U
\text{ .}
\end{equation*}
By augmenting $\Delta_\Pi$ with $\Lambda_\Pi$,
Theorem~\ref{thm:nogoods:tight} can be extended to 
non-tight programs.
%
\begin{theorem}[\cite{gekanesc07a}]\label{thm:nogoods:nontight}
Let $\Pi$ be a logic program.
Then, $X\subseteq\atom{\Pi}$ is an answer set of~$\Pi$
iff
$X=\Tass{A}\cap\atom{\Pi}$ for a (unique) 
solution $A$ for $\Delta_\Pi\cup\Lambda_\Pi$.
\end{theorem}

By virtue of Theorem~\ref{thm:nogoods:nontight},
the nogoods in $\Delta_\Pi\cup\Lambda_\Pi$ provide us with a
constraint-based characterization of the \emph{answer sets} of~$\Pi$.
However, it is important to note that the size of $\Delta_\Pi$ is linear in
$\atom{\Pi}{\times}\body{\Pi}$, while $\Lambda_\Pi$ contains exponentially many nogoods.
As shown in~\cite{lifraz04a}, under current assumptions in complexity theory,
the exponential number of elements in $\Lambda_\Pi$ is inherent, that is,
it cannot be reduced significantly in the worst case.
Hence, ASP solvers do not determine the nogoods in~$\Lambda_\Pi$ a priori,
but include mechanisms to determine them on demand.
This is illustrated further in the next section.

%% file: algorithms.tex
\section{Algorithmic Background}

This section recalls the basic decision procedure of \clasp\ \cite{gekanesc07a},
abstracting Conflict-Driven Clause Learning (CDCL; \cite{mitchell05a}) for SAT solving
from clauses, that is, Conflict-Driven Nogood Learning (CDNL).

\subsection{Conflict-Driven Nogood Learning}

Algorithm~\ref{algo:cdnl} shows our main procedure for deciding
whether a program~$\Pi$ has some answer set.
The algorithm starts with an empty assignment~$A$ and an empty set~$\nabla$
of recorded nogoods (Lines~1--2).
Note that dynamic nogoods added to~$\nabla$ in Line~5 are elements of
$\Lambda_\Pi$, while those added in Line~9 result from conflict analysis (Line~8).
In addition to conflict-driven learning, the procedure performs backjumping
(Lines~10--11), guided by a decision level~$k$ determined by conflict analysis.
Via decision level~$\mathit{dl}$, we count \emph{decision literals},
that is, literals in $A$ that have been heuristically selected in Line~15.
The initial value of $\mathit{dl}$ is~$0$ (Line~3), 
and it is incremented in Line~16 before a decision literal is added to~$A$ (Line~17).
All literals in~$A$ that are not decision literals have been derived by 
propagation in Line~5, and we call them \emph{implied literals}.
For any literal~$\sigma$ in~$A$, we write $\mathit{dl}(\sigma)$ to refer
to the decision level of~$\sigma$, that is, the value $\mathit{dl}$ had when
$\sigma$ was added to $A$.
After propagation, the main loop (Lines~4--17) distinguishes three cases:
a conflict detected via a violated nogood (Lines~6--11),
a solution (Lines~12--13), or
a heuristic selection with respect to a partial assignment (Lines~14--17).
Finally, note that a conflict at decision level~$0$ signals 
that~$\Pi$ has no answer set (Line~7).
\input{Algorithms/cdcl}

\subsection{Propagation}

Our propagation procedure, shown in Algorithm~\ref{algo:up},
derives implied literals and adds them to~$A$.
Lines 3--9 describe unit propagation (cf.\
\cite{mitchell05a}) on $\Delta_\Pi \cup \nabla$.
If a conflict is detected in Line~4, unit propagation terminates immediately (Line~5).
Otherwise, in Line~6, we determine all nogoods~$\delta$ that are \emph{unit-resulting} wrt~$A$,
that is, the complement~$\overline{\sigma}$ of some literal $\sigma\in\delta$ 
must be added to~$A$ because all other literals of~$\delta$ are already true in~$A$.
If there is some unit-resulting nogood~$\delta$ (Line~7),
$A$ is augmented with~$\overline{\sigma}$ in Line~8.
Observe that~$\delta$ is chosen non-deterministically,
and several distinct nogoods may imply $\overline{\sigma}$ wrt~$A$.
This non-determinism gives rise to our study of heuristics for conflict resolution,
selecting a resolvent among the nogoods~$\delta$ that imply~$\overline{\sigma}$.

The second part of Algorithm~\ref{algo:up} (Lines~10--14)
checks for unit-resulting or violated nogoods in~$\Lambda_\Pi$.
If~$\Pi$ is tight (Line~10), sophisticated checks are unnecessary
(cf.\ Theorem~\ref{thm:nogoods:tight}).
Otherwise, we consider sets $U\subseteq\atom{\Pi}$ such that $\EB{U}{\Pi}\subseteq\Fass{A}$,
called \emph{unfounded sets} \cite{gerosc91a}.
An unfounded set~$U$ is determined in Line~12 by a dedicated algorithm,
where $U\cap\Fass{A}=\emptyset$.
If such a nonempty unfounded set~$U$ exists, each nogood $\delta\in\Lambda_U$ is
either unit-resulting or violated wrt~$A$, and an arbitrary $\delta\in\Lambda_U$
is recorded in Line~14 for triggering unit propagation.
Note that all atoms in~$U$ must be falsified before
another unfounded set is determined (cf.\ Lines~11--12).
Eventually, propagation terminates in Line~13 if
no nonempty unfounded set has been detected in Line~12.
%
\input{Algorithms/up}

\input{Algorithms/conf}

\subsection{Conflict Analysis}

Algorithm \ref{algo:conf} shows our conflict analysis procedure,
which is based on resolution.
Given a nogood~$\delta$ that is violated wrt~$A$,
we determine in Line~2 the literal~$\sigma\in\delta$ added last to~$A$.
If~$\sigma$ is the single literal of its decision level~$\mathit{dl}(\sigma)$ in~$\delta$ (cf.\ Line~3),
it is called a \emph{unique implication point} (UIP; \cite{marsak99a}).
Among a number of conflict resolution schemes, the 
\emph{First-UIP} scheme, stopping conflict resolution as soon as the
first UIP is reached, has turned out to be the most efficient and 
most robust strategy \cite{zamamoma01a}.
Our conflict analysis procedure follows the First-UIP scheme by performing conflict resolution
only if~$\sigma$ is not a UIP (tested in Line~4) and, otherwise,
returning~$\delta$ along with the smallest decision level~$k$ at which
$\overline{\sigma}$ is implied by~$\delta$ after backjumping (Line~8).

Let us take a closer look at conflict resolution steps in Lines~5--7.
It is important to note that, if~$\sigma$ is not a UIP, it cannot be the
decision literal of~$\mathit{dl}(\sigma)$.
Rather, it must have been implied by some nogood $\varepsilon\in\Delta_\Pi\cup\nabla$.
As a consequence, the set~$\Sigma$ determined in Line~5 cannot be empty,
and we call its elements \emph{antecedents} of~$\sigma$.
Note that each antecedent~$\varepsilon$ contains~$\overline{\sigma}$ and had
been unit-resulting immediately before~$\sigma$ was added to~$A$;
we thus call $\varepsilon\setminus\{\overline{\sigma}\}$ a \emph{reason} for $\sigma$.
Knowing that~$\sigma$ may have more than one antecedent,
a non-deterministic choice among them is made in Line~6.
Exactly this choice is subject to the heuristics studied below.
Furthermore, as~$\sigma$ is the literal of~$\delta$ added last to~$A$,
$\delta\setminus\{\sigma\}$ is also a reason for $\overline{\sigma}$.
Since they imply complementary literals, no solution can jointly contain both reasons, viz.,
$\delta\setminus\{\sigma\}$ and $\varepsilon\setminus\{\overline{\sigma}\}$.
Hence, combining them in Line~7 gives again a nogood violated wrt~$A$.
Finally, note that conflict resolution is guaranteed to terminate
at some UIP, but different heuristic choices in Line~6 may result in different UIPs.



%% file: Algorithms/cdcl.tex
\begin{algorithm}[t]
\Input{A program $\Pi$.}
\Output{An answer set of $\Pi$.}
\BlankLine
$A\leftarrow\emptyset$\hfill\textit{// assignment over $\atom{\Pi}\cup\body{\Pi}$}\;
$\nabla\leftarrow\emptyset$\hfill\textit{// set of (dynamic) nogoods}\;
$\mathit{dl}\leftarrow 0$\hfill\textit{// decision level}\;
\Loop{}{
  $(A,\nabla)\leftarrow\Propagation{$\Pi,\nabla,A$}$\;
  \uIf{$\varepsilon\subseteq A \textnormal{ for some } \varepsilon\in\Delta_\Pi\cup\nabla$}
  {\lIf{$\mathit{dl}=0$}{\Return no answer set}\;
   
$(\delta,k)\leftarrow\Analysis{$\varepsilon,\Pi,\nabla,A$}$\;
    $\nabla\leftarrow \nabla\cup\{\delta\}$\;
    $A\leftarrow A\setminus\{\sigma\in A \mid k < \mathit{dl}(\sigma)\}$\;
    $\mathit{dl}\leftarrow k$
  }
  \uElseIf{$\Tass{A}\cup\Fass{A}=\atom{\Pi}\cup\body{\Pi}$}
  {\Return $\Tass{A}\cap\atom{\Pi}$}
  \Else{%
    $\sigma_d\leftarrow\Select{$\Pi,\nabla,A$}$\;
    $\mathit{dl}\leftarrow \mathit{dl}+1$\;
    $A\leftarrow A\circ(\sigma_d)$
  }
}
\caption{\textsc{CDNL}\label{algo:cdnl}}
\end{algorithm}


%% file: Algorithms/up.tex
\begin{algorithm}[t]
\Input{A program $\Pi$, a set $\nabla$ of nogoods, and an assignment $A$.}
\Output{An extended assignment and set of nogoods.}
\BlankLine
$U \leftarrow \emptyset$\hfill\textit{// unfounded set}\;
\Loop{}{
  \Repeat{$\Sigma=\emptyset$}{
    \If{$\delta\subseteq A \textnormal{ for some } \delta\in\Delta_\Pi\cup\nabla$}{\Return $(A,\nabla)$\;}
    $\Sigma \leftarrow \{\delta\in\Delta_\Pi\cup\nabla \mid \delta\setminus A =\{\sigma\},\overline{\sigma}\notin A \}$\;
    \lIf{$\Sigma\neq\emptyset$}{%
      \Let{$\sigma\in\delta\setminus A \textnormal{ for some } \delta\in\Sigma$}{$A \leftarrow A \circ(\overline{\sigma})$}
    }
  }
\BlankLine
  \lIf{
       $\textsc{Tight}(\Pi)$}{%
    \Return $(A,\nabla)$\;
  }
    $U\leftarrow U\setminus\Fass{A}$\;
    \lIf{$U=\emptyset$}{%
      $U\leftarrow\UnFoundedSet{$\Pi,A$}$
    }\;
    \lIf{$U=\emptyset$}{%
      \Return $(A,\nabla)$\;
    }\;        
      \lLet{$\delta\in\Lambda_U$}{%
        $\nabla\leftarrow \nabla\cup\{\delta\}$\;
      }
}
\caption{\textsc{Propagation}\label{algo:up}}
\end{algorithm}


%% file: Algorithms/conf.tex
\begin{algorithm}[t]
\Input{A violated nogood $\delta$, a program $\Pi$, a set $\nabla$ of nogoods, and an assignment $A$.}
\Output{A derived nogood and a decision level.}
\BlankLine
\Loop{}{
\lSuch{$\sigmaUIP\in\delta$}{%
   $\delta\setminus A[\sigmaUIP]=\{\sigmaUIP\}$
}\;
$k\leftarrow\mathit{max}(\{\mathit{dl}(\sigmaAUX)\mid\sigmaAUX\in\delta\setminus\{\sigmaUIP\}\}\cup\{0\})$\;
\uIf{$k=\mathit{dl}(\sigmaUIP)$}{%
$\Sigma \leftarrow \{\varepsilon\in\Delta_\Pi\cup\nabla \mid
\varepsilon\setminus A[\sigmaUIP]=\{\overline{\sigmaUIP}\}\}$\;
  $\varepsilon \leftarrow \SelectAntecedent{$\Sigma$}$\;
$\delta\leftarrow(\delta\setminus\{\sigmaUIP\}) \cup (\varepsilon \setminus\{
\overline{\sigmaUIP}\})$
}
\lElse{\Return $(\delta,k)$}
}
\caption{\textsc{ConflictAnalysis}\label{algo:conf}}
\end{algorithm}


%% file: cnflanalysis.tex
\section{Implication Graphs and Conflict Graphs}

To portray the matter of choosing among several distinct antecedents,
we modify the notion of an implication graph \cite{bekasa04a}.
At a given state of CDNL,
the \emph{implication graph} contains a node for each literal~$\sigma$ in assignment~$A$
and, for a violated nogood $\delta\subseteq A$,
a node~$\overline{\sigma}$ is included,
where~$\sigma$ is the literal of~$\delta$ added last to~$A$,
that is, $\delta\setminus A[\sigma]=\{\sigma\}$.
Furthermore, for each antecedent~$\delta$ of an implied literal~$\sigma$,
the implication graph contains directed edges labeled with~$\delta$ 
from all literals in the reason $\delta\setminus\{\overline{\sigma}\}$ to $\sigma$.
Different from \cite{bekasa04a},
where implication graphs reflect exactly one reason per implied literal,
our implication graph thus includes all of them.
If the implication graph contains both~$\sigma$ and~$\overline{\sigma}$,
we call them \emph{conflicting literals}. 
Note that an implication graph contains at most one such pair 
$\{\sigma,\overline{\sigma}\}$, called \emph{conflicting assignment},
because our propagation procedure in Algorithm~\ref{algo:up} stops
as soon as a nogood becomes violated (cf.\ Lines~4--5).

\begin{figure}
\centering
\includegraphics[width=\linewidth]{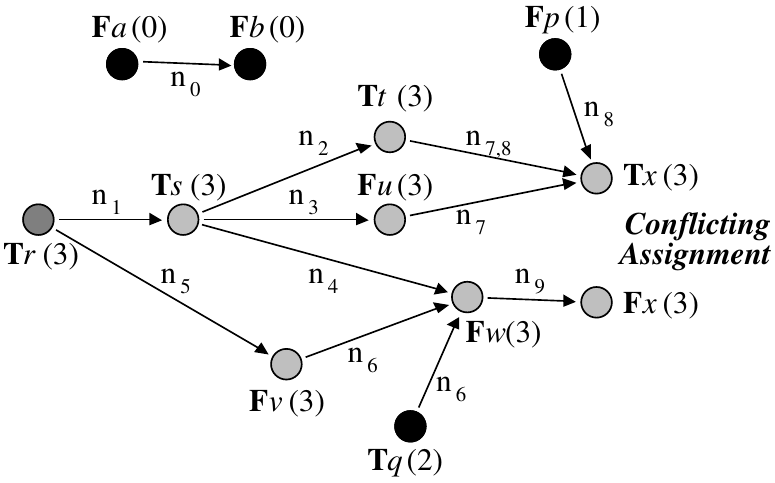}
\caption{An exemplary implication graph containing a conflicting assignment.}
\label{fig:implication}
\end{figure}

An exemplary implication graph is shown in Figure~\ref{fig:implication}.
Each of its nodes (except for one among the two conflicting literals) 
corresponds to a literal that is true in assignment
\begin{equation*}
A = 
\big(
 \Flit{a}, \Flit{b}, \underline{\Flit{p}}, \underline{\Tlit{q}}, \underline{\Tlit{r}},
 \Tlit{s}, \Flit{v}, \Tlit{t}, \Flit{u}, \Flit{w}, \Tlit{x}
\big)
\ \text{.}
\end{equation*}
The three decision literals in~$A$ are underlined, and all other literals are implied.
For each literal~$\sigma$, its decision level~$\mathit{dl}(\sigma)$ is also provided
in Figure~\ref{fig:implication} in parentheses.
Every edge is labeled with at least one antecedent of its target,
that is, the edges represent the following nogoods:
\begin{align*}
n_0 &= \{\Flit{a}, \Tlit{b}\} & n_1 &= \{\Tlit{r}, \Flit{s}\} \\
n_2 &= \{\Tlit{s}, \Flit{t}\} & n_3 &= \{\Tlit{s}, \Tlit{u}\} \\
n_4 &= \{\Tlit{s}, \Tlit{w}\} & n_5 &= \{\Tlit{r}, \Tlit{v}\} \\
n_6 &= \{\Tlit{q}, \Flit{v}, \Tlit{w}\} &
n_7 &= \{\Tlit{t}, \Flit{u}, \Flit{x}\} \\
n_8 &= \{\Flit{p}, \Tlit{t}, \Flit{x}\} &
n_9 &= \{\Flit{w}, \Tlit{x}\} \ \text{.}
\end{align*}
Furthermore, nogood $\{\Tlit{a}\}$ is unit-resulting wrt the empty assignment,
thus, implied literal~$\Flit{a}$ (whose decision level is~$0$) does not have any incoming edge.
Observe that the implication graph contains conflicting assignment $\{\Tlit{x}, \Flit{x}\}$,
where $\Tlit{x}$ has been implied by nogood~$n_7$ and likewise by~$n_8$.
It is also the last literal in~$A$ belonging to violated nogood~$n_9$,
so that its complement~$\Flit{x}$ is the second conflicting literal in the implication graph.
Besides~$\Tlit{x}$, literal~$\Flit{w}$ has multiple antecedents, namely, $n_4$ and~$n_6$,
which can be read off the labels of the incoming edges of~$\Flit{w}$.

The conflict resolution done in Algorithm~\ref{algo:conf},
in particular, the heuristic choice of antecedents in Line~6,
can now be viewed as an iterative projection of the implication graph.
In fact, if an implied literal has incoming edges with distinct labels,
all edges with a particular label are taken into account,
while the edges with different labels only are dropped.
This observation motivates the following definition:
a subgraph of an implication graph is a \emph{conflict graph} if it
contains a conflicting assignment and, for each implied literal~$\sigma$ in the subgraph,
the set of predecessors of~$\sigma$ is a reason for~$\sigma$.
Note that this definition allows us to drop all literals that do not have
a path to any conflicting literal,
such as $\Flit{a}$ and $\Flit{b}$ in Figure~\ref{fig:implication}.
Furthermore, the requirement that the predecessors of an implied literal form a reason
corresponds to the selection of an antecedent, where
only the incoming edges with a particular label are traced via conflict resolution.

The next definition accounts for a particularity of ASP solving related to unfounded set handling:
a conflict graph is \emph{level-aware} if each conflicting literal~$\sigma$
has some predecessor~$\sigmaAUX$ such that $\mathit{dl}(\sigmaAUX)=\mathit{dl}(\sigma)$.
In fact, propagation in Algorithm~\ref{algo:up} is limited to falsifying unfounded atoms,
thus, unit propagation on nogoods in~$\Lambda_\Pi$ is performed only partially and
may miss implied literals corresponding to external bodies (cf.\ \cite{gekanesc07a}).
If a conflict graph is not level-aware, the violated nogood~$\delta$ provided as input
to Algorithm~\ref{algo:conf} already contains a UIP, thus,
$\delta$ itself is returned without performing any conflict resolution in-between.
Given that we are interested in conflict resolution,
we below consider level-aware conflict graphs only.

\begin{figure}
 \centering
\includegraphics[width=\linewidth]{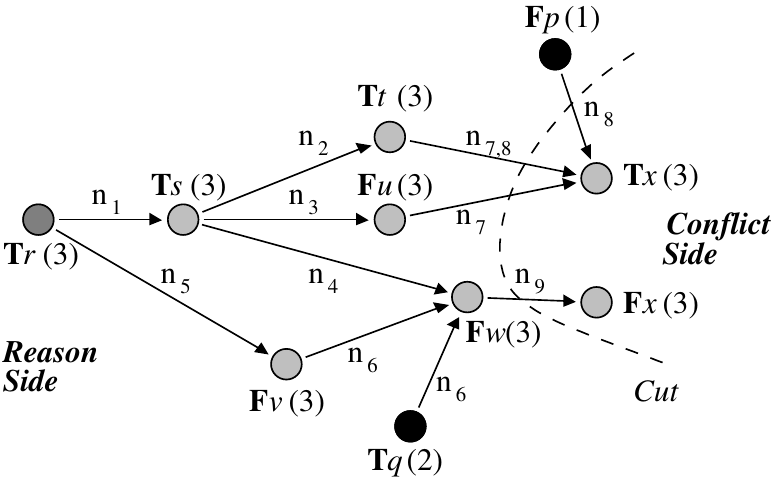}
\caption{The implication graph with a First-New-Cut.}
\label{fig:conflict00}
\end{figure}

Finally, we characterize nogoods derived by Algorithm~\ref{algo:conf}
by cuts in conflict graphs (cf.\ \cite{zamamoma01a,bekasa04a}). 
A \emph{conflict cut} in a conflict graph is a bipartition of the nodes such that
all decision literals belong to one side, called \emph{reason side}, 
and the conflicting assignment is contained in the other side, called \emph{conflict side}.
The set of nodes on the reason side that have some edge into the conflict side
form the \emph{conflict nogood} associated with a particular conflict cut.
For illustration,
a First-New-Cut \cite{bekasa04a} is shown in Figure \ref{fig:conflict00}.
For the underlying conflict graph, we can choose among the incoming edges of~$\Tlit{x}$
whether to include the edges labeled with~$n_7$ or the ones labeled with $n_8$.
With $n_7$, we get conflict nogood $\{\Tlit{t}, \Flit{u}, \Flit{w}\}$,
while $n_8$ yields $\{\Flit{p}, \Tlit{t}, \Flit{w}\}$.
%
%
%
%
%

Different conflict cuts correspond to different resolution schemes,
where we are particularly interested in the First-UIP scheme.
Given a conflict graph and conflicting assignment $\{\sigma,\overline{\sigma}\}$,
a UIP~$\sigma_{\mathit{UIP}}$ can be identified as a node such that all paths
from~$\sigma_d$, the decision literal of decision level $\mathit{dl}(\sigma)=\mathit{dl}(\overline{\sigma})$,
to either~$\sigma$ or~$\overline{\sigma}$ go through~$\sigma_{\mathit{UIP}}$
(cf.\ \cite{zamamoma01a}).
In view of this alternative definition of a UIP,
it becomes even more obvious than before that~$\sigma_d$ is indeed a UIP, also called the Last-UIP.
In contrast, a literal~$\sigma_{\mathit{UIP}}$ is the First-UIP if it is the UIP ``closest''
to the conflicting literals, that is, if no other UIP is reachable from~$\sigma_{\mathit{UIP}}$.
The \emph{First-UIP-Cut} is then given by the conflict cut that has all literals
lying on some path from the First-UIP to a conflicting literal, except for the First-UIP itself,
on the conflict side and all other literals (including the First-UIP) on the reason side.
The \emph{First-UIP-Nogood}, that is, the conflict nogood associated with the First-UIP-Cut,
is exactly the nogood derived by conflict resolution in Algorithm~\ref{algo:conf}
when antecedents that contribute edges to the conflict graph are selected for conflict resolution.
Also note that the First-UIP-Cut for a conflict graph is unique, thus,
by projecting an implication graph to a conflict graph,
we implicitly fix the First-UIP-Nogood.
With this is mind,
the next section deals with heuristics for extracting conflict graphs from implication graphs.



%% file: heuristics.tex
\section{Heuristics}

In this section, we propose several heuristics for conflict resolution
striving for different goals.

\subsection{Recording Short Nogoods}

\begin{figure}
 \centering
\includegraphics[width=\linewidth]{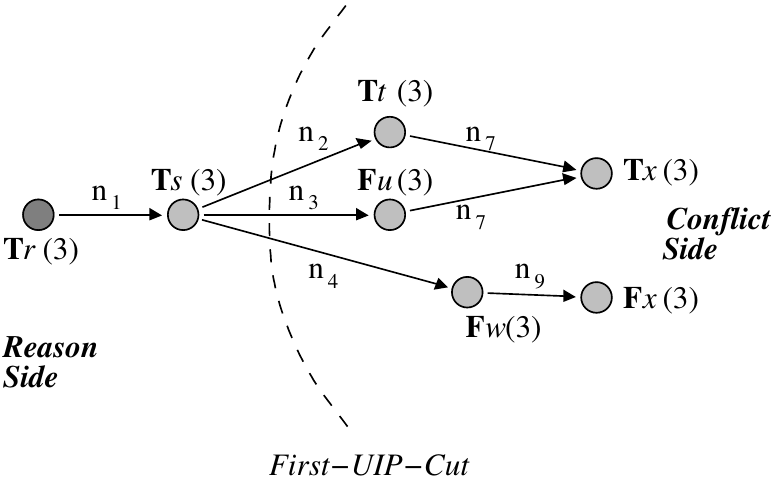}
\caption{A First-UIP-Cut obtained with $H_{\mathit{short}}$.}
\label{fig:conf02}
\end{figure}

Under the assumption that short nogoods prune larger portions of the search 
space than longer ones, a First-UIP-Nogood looks the more attractive the
less literals it contains.
In addition, unit propagation on shorter nogoods is usually faster
and might even be enabled to use particularly optimized data structures,
for instance, specialized to binary or ternary nogoods \cite{ryan04a}.
As noticed in \cite{mafuma04a},
a conflict nogood stays short when the resolvents are short,
when the number of resolvents is small,
or when the resolvents have many literals in common.
In the SAT area,
it has been observed that preferring short nogoods in conflict resolution
may lead to resolution sequences involving mostly binary and ternary nogoods,
so that derived conflict nogoods are not much longer
than the originally violated nogoods \cite{mitchell05a}.
Our first heuristics, $H_{\mathit{short}}$, thus selects an antecedent
containing the smallest number of literals among the available antecedents of a literal.
Given the same implication graph as in Figure~\ref{fig:implication} and~\ref{fig:conflict00},
$H_{\mathit{short}}$ may yield the conflict graph shown in Figure~\ref{fig:conf02}
by preferring antecedent~$n_7$ of~$\Tlit{x}$ over~$n_8$ and
antecedent~$n_4$ of~$\Flit{w}$ over~$n_6$ during conflict resolution.
The corresponding First-UIP-Nogood, $\{\Tlit{s}\}$, is indeed short and
enables CDNL to after backjumping derive $\Flit{s}$ by unit propagation
at decision level~$0$.
However, the antecedents~$n_7$ and~$n_8$ of~$\Tlit{x}$ are of the same size,
thus, $H_{\mathit{short}}$ may likewise pick~$n_8$,
in which case the First-UIP-Cut in Figure~\ref{fig:conf01} is obtained.
The corresponding First-UIP-Nogood, $\{\Flit{p},\Tlit{s}\}$, is longer.
Nonetheless, our experiments below empirically confirm that $H_{\mathit{short}}$
tends to reduce the size of First-UIP-Nogoods.
But before, we describe further heuristics focusing also on other aspects.%

\subsection{Performing Long Backjumps}

\begin{figure}
\centering
\includegraphics[width=\linewidth]{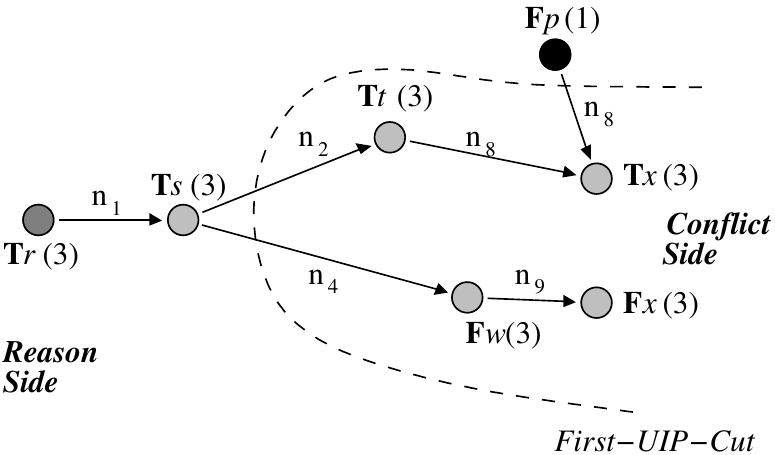}
\caption{A First-UIP-Cut obtained with $H_{\mathit{lex}}$.}
\label{fig:conf01}
\end{figure}

By backjumping, 
CDNL may skip the exhaustive exploration of regions of the search space,
possibly escaping spare regions not containing any solution.
Thus, it seems reasonable to aim at First-UIP-Nogoods such that
their literals belong to small decision levels,
as they are the determining factor for the lengths of backjumps.
Our second heuristics, $H_{\mathit{lex}}$, thus uses a 
lexicographic order to rank antecedents according to the
decision levels of their literals.
Given an antecedent~$\delta$ of a literal~$\sigma$,
we arrange the literals in the reason 
$\delta\setminus\{\overline{\sigma}\}$ for~$\sigma$ in
descending order of their decision levels.
The so obtained sequence $(\sigma_1,\dots,\sigma_m)$, where 
$\delta\setminus\{\overline{\sigma}\}=\{\sigma_1,\dots,\sigma_m\}$,
induces a descending list 
$\mathit{levels}(\delta)=(\mathit{dl}(\sigma_1),\dots,\mathit{dl}(\sigma_m))$
of decision levels.
An antecedent~$\delta$ is then considered to be smaller 
than another antecedent~$\varepsilon$, viz., $\delta < \varepsilon$, if
the first element that differs in $\mathit{levels}(\delta)$ and
$\mathit{levels}(\varepsilon)$ is smaller in $\mathit{levels}(\delta)$
or if
$\mathit{levels}(\delta)$ is a prefix of $\mathit{levels}(\varepsilon)$ and
shorter than $\mathit{levels}(\varepsilon)$.
Due to the last condition,
$H_{\mathit{lex}}$ also prefers an antecedent~$\delta$ that is shorter than~$\varepsilon$,
provided that literals of the same decision levels as in~$\delta$
are also found in~$\varepsilon$.
Reconsidering the implication graph in
Figure~\ref{fig:implication} and~\ref{fig:conflict00},
we obtain
$\mathit{levels}(n_8)=(3,1)<(3,3)=\mathit{levels}(n_7)$
for antecedents~$n_7$ and~$n_8$ of~$\Tlit{x}$, and we have
$\mathit{levels}(n_4)=(3)<(3,2)=\mathit{levels}(n_6)$
for antecedents~$n_4$ and~$n_6$ of~$\Flit{w}$.
By selecting antecedents that are lexicographically smallest,
$H_{\mathit{lex}}$ leads us to the conflict graph shown in Figure~\ref{fig:conf01}.
In this example,
the corresponding First-UIP-Nogood, $\{\Flit{p},\Tlit{s}\}$, 
is weaker than $\{\Tlit{s}\}$, which may be obtained with $H_{\mathit{short}}$
(cf.\ Figure~\ref{fig:conf02}).

\begin{figure}
\centering
\includegraphics[width=\linewidth]{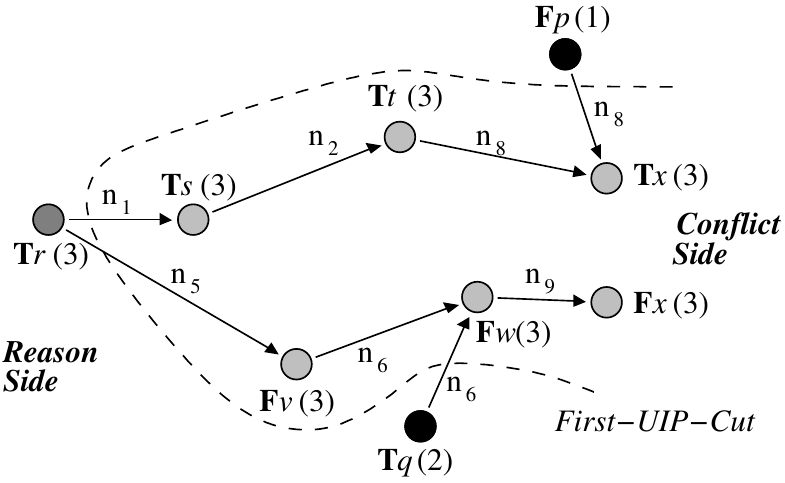}
\caption{A First-UIP-Cut obtained with $H_{\mathit{avg}}$.}
\label{fig:conf03}
\end{figure}

Given that lexicographic comparisons are computationally expensive,
we also consider a lightweight variant of ranking antecedents
according to decision levels.
Our third heuristics, $H_{\mathit{avg}}$, 
prefers an antecedent~$\delta$ over~$\varepsilon$ if the
average of $\mathit{levels}(\delta)$ is smaller than the
average of $\mathit{levels}(\varepsilon)$.
In our example,
we get 
$\mathit{avg}[\mathit{levels}(n_8)] =
 \mathit{avg}(3,1) = 2 < 3 = 
 \mathit{avg}(3,3) =
 \mathit{avg}[\mathit{levels}(n_7)]$
and
$\mathit{avg}[\mathit{levels}(n_6)] =
 \mathit{avg}(3,2) = 2.5 < 3 = 
 \mathit{avg}(3) =
 \mathit{avg}[\mathit{levels}(n_4)]$,
yielding the conflict graph shown in Figure~\ref{fig:conf03}.
Unfortunately,
the corresponding First-UIP-Nogood, $\{\Flit{p},\Tlit{q},\Tlit{r}\}$,
does not match the goal of $H_{\mathit{avg}}$ as
backjumping only returns to decision level~$2$,
where~$\Tlit{r}$ is then flipped to~$\Flit{r}$.
Note that this behavior is similar to chronological backtracking,
which can be regarded as the most trivial form of backjumping.
%

\subsection{Shortening Conflict Resolution}

Our fourth heuristics, $H_{\mathit{res}}$,
aims at speeding up conflict resolution itself by shortening resolution sequences.
In order to earlier encounter a UIP,
$H_{\mathit{res}}$ prefers antecedents such that the
number of literals at the current decision level~$\mathit{dl}$ is smallest.
%
In our running example, $H_{\mathit{res}}$ prefers~$n_8$ over~$n_7$ as it
contains fewer literals whose decision level is~$3$.
However, antecedents~$n_4$ and~$n_6$ of~$\Flit{w}$ are indifferent, thus,
$H_{\mathit{res}}$ may yield either one of the conflict graphs in Figure~\ref{fig:conf01}
and~\ref{fig:conf03}.

\subsection{Search Space Pruning}

The heuristics presented above
rank antecedents merely by structural properties,
thus disregarding their contribution in the past to 
solving the actual problem.
The latter is estimated by nogood deletion heuristics of SAT solvers
\cite{golnov02a,mafuma04a},
and \clasp\ also maintains activity scores for nogoods
\cite{gekanesc07b}.
Our fifth heuristics, $H_{\mathit{active}}$,
makes use of them and ranks antecedents according to their activities.

Finally,
we investigate a heuristics, $H_{\mathit{prop}}$,
that stores (and prefers) the smallest decision level at which a nogood has ever been unit-resulting.
%
The intuition underlying $H_{\mathit{prop}}$ is that the number of implied literals
at small decision levels can be viewed as a measure for the progress of CDNL,
in particular, as attesting unsatisfiability requires a conflict at decision level~$0$.
Thus, it might be a good idea to prefer nogoods that gave rise to implications
at small decision levels.


%% file: experiments.tex
\section{Experiments} 

For their empirical assessment,
we have implemented the heuristics proposed above in
a prototypical extension of our ASP solver \clasp\ version 1.0.2.
(Even though there are newer versions of \clasp,
 a common testbed, omitting some optimizations,
 is sufficient for a representative comparison.)
Note that \clasp\ \cite{gekanesc07b} incorporates various advanced
Boolean constraint solving techniques,
e.g.:
\begin{itemize}
\item
lookback-based decision heuristics \cite{golnov02a},
\item
restart and nogood deletion policies \cite{eensor03a},
\item
watched literals for unit propagation on ``long'' nogoods \cite{momazhzhma01a},
\item
dedicated treatment of binary and ternary nogoods \cite{ryan04a}, and
\item
early conflict detection \cite{mafuma04a}.
\end{itemize}
Due to this variety,
the solving process of \clasp\ is a complex interplay of different features.
Thus, it is almost impossible to observe the impact of a certain feature,
such as our conflict resolution heuristics, in isolation.
However, we below use a considerable number of benchmark classes with different characteristics and
shuffled instances,
so that noise effects should be compensated at large.

For accommodating conflict resolution heuristics
considering several antecedents per literal,
the low-level implementation of \clasp\ had to be modified.
These modifications are less optimized than the original implementation,
so that our prototype incurs some disadvantages in raw speed that can potentially
be reduced by optimizing the implementation.
However, for comparison, we include unmodified \clasp\ version 1.0.2,
not applying any particular heuristics in conflict resolution.
Given that unit propagation in \clasp\ privileges binary and ternary
nogoods, they are more likely to be used as antecedents than longer nogoods,
as original \clasp\ simply stores the first antecedent it encounters and ignores others.
In view of this, unit propagation of original \clasp\ leads conflict resolution
into the same direction as $H_{\mathit{short}}$, though in a less exact way.
The next table summarizes all \clasp\ variants and
conflict resolution heuristics under consideration,
denoting the unmodified version simply by \clasp:

\noindent
\begin{tabular}{@{}|c||c|l|}
\hline
Label & Heuristics & Goal
\\\hline\hline
\clasp
&
---
&
speeding up unit propagation
\\\hline
\claspsmall
&
$H_{\mathit{short}}$
&
recording short nogoods
\\\hline
\clasplex
&
$H_{\mathit{lex}}$
&
performing long backjumps
\\\hline
\claspavg
&
$H_{\mathit{avg}}$
&
performing long backjumps
\\\hline
\clasplcd
&
$H_{\mathit{res}}$
&
shortening conflict resolution
\\\hline
\claspactive
&
$H_{\mathit{active}}$
&
search space pruning
\\\hline
\claspprop
&
$H_{\mathit{prop}}$
&
search space pruning
\\\hline
\end{tabular}

\noindent
Note that all \clasp\ variants perform early conflict detection, that is,
they encounter a unique conflicting assignment before beginning with conflict resolution.
Furthermore, all of them perform conflict resolution according to the First-UIP scheme.
Thus, we do not explore the first two among the three degrees of freedom
mentioned in the introductory section and concentrate fully on the choice of resolvents. 

%
%
We conducted experiments on the benchmarks used in categories
\emph{SCore} and \emph{SLparse} 
of the first ASP system competition \cite{contest07a}.
Tables~\ref{tab:propone}--\ref{tab:one} group benchmark instances
by their classes, viz., Classes~1--11.
Via superscripts~$^s$ and~$^r$ in the first column,
we indicate whether the~$n$ instances belonging to a class are structured
(e.g., 15-Puzzle) or randomly generated (e.g., BlockedN-Queens).
We omit classifying Factoring, which is a worst-case problem where
an efficient algorithm would yield a cryptographic attack.
Furthermore, Tables~\ref{tab:propone}--\ref{tab:one} show results
for computing one answer set or deciding that an instance has no answer set.
For each benchmark instance, we performed five runs on different shuffles,
resulting in $5n$ runs per benchmark class.
All experiments were run on a 3.4GHz PC under Linux;
each run was limited to 600s time and 1GB RAM.
Note that, in Tables \ref{tab:propone}--\ref{tab:propthree},
we consider only the instances on which runs were
completed by all considered \clasp\ variants.

Table \ref{tab:propone} shows the average lengths of
First-UIP-Nogoods for the heuristics aiming at short nogoods,
implemented by \claspsmall\ and \clasplex, among which the latter uses
the lengths of antecedents as a tie breaker.
For comparison, we also include original \clasp.
On most benchmark classes,
we observe that
\claspsmall\ as well as \clasplex\ tend to reduce the lengths of First-UIP-Nogoods,
up to~$14$ percent shorter than the ones of \clasp\ on BlockedN-Queens. 
But there remains only a slight reduction of about~$6$ percent 
shorter First-UIP-Nogoods of \clasplex\ in the summary of all benchmark
classes (weighted equally).
We also observe that \claspsmall, 
more straightly preferring short antecedents than \clasplex,
does not reduce First-UIP-Nogood lengths any further.
Interestingly, there is no clear distinction between structured and randomly
generated instances, neither regarding magnitudes nor reduction rates
of First-UIP-Nogood lengths.
%
\input{Tables/propone}

Table \ref{tab:proptwo} shows the average backjump lengths 
in terms of decision levels for the \clasp\ variants aiming at long backjumps,
viz., \claspavg\ and \clasplex.
We note that average backjump lengths of more than~$2$ decision levels
indicate structured instances, except for BoundedSpanningTree.
Regarding the increase of backjump lengths, \claspavg\ does not
exhibit significant improvements, and the polarity of differences 
to original \clasp\ varies.
Only the more sophisticated heuristics of \clasplex\ 
almost consistently leads to increased backjump lengths (except for HamiltonianPath),
but the amounts of improvements are rather small.
\input{Tables/proptwo}

Table \ref{tab:propthree} shows the average numbers of conflict resolution steps
for \clasplcd\ and \clasplex, among which the former particularly aims at
their reduction.
Somewhat surprisingly, \clasplcd\ in all performs more conflict resolution steps
even than original \clasp,
while \clasplex\ almost consistently exhibits a reduction of conflict resolution steps
(except for Su-Doku).
This negative result for \clasplcd\ suggests that trimming
conflict resolution regardless of its outcome is not advisable.
The quality of recorded nogoods certainly is a key factor for the
performance of conflict-driven learning solvers for ASP and SAT,
thus, shallow savings in their retrieval are not worth it and might
even be counterproductive globally.
%
%
%
\input{Tables/propthree}

Finally, Table \ref{tab:one} provides average numbers of conflicts and
average runtimes in seconds for all \clasp\ variants.
For each benchmark class, the first line provides the average numbers of
conflicts encountered on instances where runs were completed by all \clasp\ variants,
while the second line gives the average times of completed runs 
and numbers of timeouts in parentheses.
(Recall that all \clasp\ variants were run on~$5n$ shuffles of the~$n$
 instances per class, leading to more than~$n$ timeouts on BlockedN-Queens
 and, with some \clasp\ variants, also on Solitaire.)
At the bottom of~Table \ref{tab:one},
we summarize average numbers of conflicts and average runtimes
over all benchmark classes (weighted equally).
Note that the last but one line provides the sums of timeouts in parentheses, 
while the last line penalizes timeouts with maximum time, viz., 600 seconds.
%
As mentioned above, original \clasp\ is highly optimized and does not suffer
from the overhead incurred by the extended infrastructure for applying
heuristics in conflict resolution.
As a consequence,
we observe that original \clasp\ outperforms its variants
on most benchmark classes as regards runtime.
Among the variants of \clasp,
\claspavg\ in all exhibits the best average number of conflicts and runtime.
However, it also times out most often and behaves unstable,
as the poor performance on Classes~2 and~11 shows.
In contrast, \claspsmall\ and \clasplex\ lead to fewest timeouts 
(in fact, as many timeouts as \clasp),
and \clasplex\ encounters fewer conflicts than \claspsmall.
Variant \claspactive, preferring ``critical'' antecedents,
exhibits a comparable performance,
while \clasplcd\ and \claspprop\ yield more timeouts and
also encounter relatively many conflicts.
Overall, we notice that some \clasp\ variants perform reasonably well,
but without significantly decreasing the number of conflicts 
in comparison to original \clasp.
As there is no clear winner among our \clasp\ variants,
unfortunately, they do not suggest any ``universal'' conflict resolution heuristics.
%
\input{Tables/one}


%% file: Tables/propone.tex
\begin{table}[t]
\centering
{\scriptsize
\begin{tabular*}{\linewidth}{|c|@{\extracolsep{\fill}}l|c||c|c||c|}
\hline 
No. & Class & $n$ & \claspsmall & \clasplex & \clasp \\
\hline\hline
$1^s$ & 15-Puzzle & $10$
& 22.33                            
& 22.35                            
& 23.03                            
\\\hline
$2^r$ & BlockedN-Queens & $7$
& 27.32                            
& 28.23                            
& 31.85                            
\\\hline
$3^s$ & EqTest & $5$
& 172.12                            
& 178.27                            
& 189.12                            
\\\hline
$4$ & Factoring & $5$
& 134.95                            
& 130.67                            
& 141.34                            
\\\hline
$5^s$ & HamiltonianPath & $14$
& 12.96                            
& 11.73                            
& 12.04                            
\\\hline
$6^r$ & RandomNonTight & $14$
& 31.82                            
& 32.07                            
& 32.74                            
\\\hline
$7^r$ & BoundedSpanningTree & $5$
& 35.06                            
& 36.68                            
& 33.95                            
\\\hline
$8^s$ & Solitaire & $4$
& 24.55                            
& 22.02                            
& 25.03                            
\\\hline
$9^s$ & Su-Doku & $3$
& 16.22                            
& 15.09                            
& 13.99                            
\\\hline
$10^s$ & TowersOfHanoi & $5$
& 52.89                            
& 52.31                            
& 58.29                            
\\\hline
$11^r$ & TravelingSalesperson & $5$
& 101.37                           
&  90.35                           
&  99.26                           
\\\hline\hline
\multicolumn{3}{|c||}{Average First-UIP-Nogood Length}
& 45.15                            
& 44.46                            
& 47.21                            
\\\hline
\end{tabular*}}
\caption{Average lengths of First-UIP-Nogoods per conflict.}
\label{tab:propone}
\end{table}

%% file: Tables/proptwo.tex
\begin{table}[t]
\centering
{\scriptsize
\begin{tabular*}{\linewidth}{|c|@{\extracolsep{\fill}}l|c||c|c||c|}
\hline 
No. & Class & $n$ & \claspavg & \clasplex & \clasp \\
\hline\hline
$1^s$ & 15-Puzzle & $10$
& 2.12                            
& 2.14                            
& 2.10                            
\\\hline
$2^r$ & BlockedN-Queens & $7$
& 1.07                            
& 1.08                            
& 1.07                            
\\\hline
$3^s$ & EqTest & $5$
& 1.03                            
& 1.04                            
& 1.03                            
\\\hline
$4$ & Factoring & $5$
& 1.20                            
& 1.21                            
& 1.20                            
\\\hline
$5^s$ & HamiltonianPath & $14$
& 2.53                            
& 2.58                            
& 2.62                            
\\\hline
$6^r$ & RandomNonTight & $14$
& 1.15                            
& 1.16                            
& 1.15                            
\\\hline
$7^r$ & BoundedSpanningTree & $5$
& 3.12                            
& 3.47                            
& 3.06                            
\\\hline
$8^s$ & Solitaire & $4$
& 3.34                            
& 3.28                            
& 2.92                            
\\\hline
$9^s$ & Su-Doku & $3$
& 2.55                            
& 3.01                            
& 2.76                            
\\\hline
$10^s$ & TowersOfHanoi & $5$
& 1.46                            
& 1.46                            
& 1.40                            
\\\hline
$11^r$ & TravelingSalesperson & $5$
& 1.27                            
& 1.51                            
& 1.43                            
\\\hline\hline
\multicolumn{3}{|c||}{Average Backjump Length}
& 1.89                            
& 1.99                            
& 1.89                            
\\\hline
\end{tabular*}}
\caption{Average backjump lengths per conflict.}
\label{tab:proptwo}
\end{table}

%% file: Tables/propthree.tex
\begin{table}[t]
\centering
{\scriptsize
\begin{tabular*}{\linewidth}{|c|@{\extracolsep{\fill}}l|c||c|c||c|}
\hline 
No. & Class & $n$ & \clasplcd & \clasplex & \clasp \\
\hline\hline
$1^s$ & 15-Puzzle & $10$
& 102.95                            
& 103.45                          
& 103.77                            
\\\hline
$2^r$ & BlockedN-Queens & $7$
& 18.17                            
& 17.61                           
& 17.74                            
\\\hline
$3^s$ & EqTest & $5$
& 86.94                            
& 84.78                           
& 85.76                            
\\\hline
$4$ & Factoring & $5$
& 325.54                            
& 290.36                          
& 296.07                            
\\\hline
$5^s$ & HamiltonianPath & $14$
& 11.87                            
& 12.03                           
& 12.14                            
\\\hline
$6^r$ & RandomNonTight & $14$
& 16.41                            
& 16.47                           
& 32.74                            
\\\hline
$7^r$ & BoundedSpanningTree & $5$
& 20.11                           
& 20.27                           
& 20.66                           
\\\hline
$8^s$ & Solitaire & $4$
& 79.05                           
& 67.70                           
& 79.89                           
\\\hline
$9^s$ & Su-Doku & $3$
& 21.48                           
& 20.86                           
& 19.73                           
\\\hline
$10^s$ & TowersOfHanoi & $5$
& 41.60                           
& 40.36                           
& 42.69                           
\\\hline
$11^r$ & TravelingSalesperson & $5$
& 141.68                          
&  96.06                          
& 122.98                          
\\\hline\hline
\multicolumn{3}{|c||}{Average Number of Resolution Steps}
& 78.71                           
& 70.00                           
& 75.83                           
\\\hline
\end{tabular*}}
\caption{Average numbers of resolution steps per conflict.}
\label{tab:propthree}
\end{table}

%% file: Tables/one.tex
\begin{table*}[t]
\centering
{\scriptsize
\begin{tabular}{|c|l|c||c|c|c|c|c|c||c|}
\hline 
No. & Class & $n$ & \claspsmall & \clasplex & \claspavg & \clasplcd
& \claspactive & \claspprop & \clasp\\
\hline\hline
\input{Tables/tabone}
\hline
\end{tabular}}
\caption{Average numbers of conflicts and runtimes.}
\label{tab:one}
\end{table*}


%% file: Tables/tabone.tex
\raisebox{-5pt}[0pt][0pt]{$1^s$} & \raisebox{-5pt}[0pt][0pt]{15-Puzzle} &
\raisebox{-5pt}[0pt][0pt]{$10$}
& 195.00                         
& 203.96                         
& 203.54                         
& 248.00                         
& 261.44                         
& 226.96                         
& 241.18                         
\\\cline{4-10}
 & &
& 0.13                      
& 0.14                      
& 0.14                      
& 0.15                      
& 0.16                      
& 0.15                      
& 0.14                      
\\\hline
\raisebox{-5pt}[0pt][0pt]{$2^r$} & \raisebox{-5pt}[0pt][0pt]{BlockedN-Queens} &
\raisebox{-5pt}[0pt][0pt]{$7$}
& 27289.06                         
& 26989.57                         
& 28176.00                         
& 27553.63                         
& 30240.71                         
& 29119.60                         
& 28588.34                         
\\\cline{4-10}
 & &
& 116.87 (24)                   
& 122.04 (21)                   
& 39.70 (27)                    
& 86.24 (24)                    
& 138.01 (22)                   
& 68.10 (25)                    
& 24.52 (22)                    
\\\hline
\raisebox{-5pt}[0pt][0pt]{$3^s$} & \raisebox{-5pt}[0pt][0pt]{EqTest}   &
\raisebox{-5pt}[0pt][0pt]{$5$}
& 62430.92                         
& 62648.96                         
& 59330.52                         
& 62705.00                         
& 62374.84                         
& 63303.44                         
& 62290.76                         
\\\cline{4-10}
 & &
& 19.47                     
& 21.66                     
& 19.41                     
& 19.98                     
& 20.03                     
& 21.30                     
& 15.66                     
\\\hline
\raisebox{-5pt}[0pt][0pt]{$4$} & \raisebox{-5pt}[0pt][0pt]{Factoring} &
\raisebox{-5pt}[0pt][0pt]{$5$}
& 15468.44                         
& 14838.64                         
& 14985.72                         
& 16016.56                         
& 16365.52                         
& 15404.64                         
& 16920.68                         
\\\cline{4-10}
 & &
& 6.30                      
& 5.85                      
& 6.27                      
& 6.55                      
& 6.36                      
& 6.36                      
& 5.11                      
\\\hline
\raisebox{-5pt}[0pt][0pt]{$5^s$} & \raisebox{-5pt}[0pt][0pt]{HamiltonianPath} &
\raisebox{-5pt}[0pt][0pt]{$14$}
& 703.70                         
& 683.29                         
& 653.19                         
& 564.83                         
& 764.16                         
& 694.33                         
& 650.70                         
\\\cline{4-10}
 & &
& 0.05                      
& 0.05                      
& 0.05                      
& 0.04                      
& 0.06                      
& 0.05                      
& 0.05                      
\\\hline
\raisebox{-5pt}[0pt][0pt]{$6^r$} & \raisebox{-5pt}[0pt][0pt]{RandomNonTight} &
\raisebox{-5pt}[0pt][0pt]{$14$}
& 427031.71                         
& 411024.73                         
& 402846.21                         
& 429955.23                         
& 423332.74                         
& 405476.81                         
& 406007.41                         
\\\cline{4-10}
 & &
& 53.85                     
& 55.17                     
& 51.53                     
& 54.92                     
& 53.33                     
& 52.78                     
& 41.79                     
\\\hline
\raisebox{-5pt}[0pt][0pt]{$7^r$} &
\raisebox{-5pt}[0pt][0pt]{BoundedSpanningTree} &
\raisebox{-5pt}[0pt][0pt]{$5$}
& 879.92                   
& 640.88                   
& 801.76                   
& 634.96                   
& 662.22                   
& 940.92                   
& 949.84                   
\\\cline{4-10}
 & &
& 4.51                         
& 4.37                         
& 4.38                         
& 4.36                         
& 4.27                         
& 4.98                         
& 4.42                         
\\\hline
\raisebox{-5pt}[0pt][0pt]{$8^s$} & \raisebox{-5pt}[0pt][0pt]{Solitaire} &
\raisebox{-5pt}[0pt][0pt]{$4$}
& 193.85                  
& 145.85                   
& 103.40                   
& 134.75                   
& 103.40                   
&  95.90                   
& 134.00                   
\\\cline{4-10}
 & &
& 66.14 (2)                        
& 0.22 (5)                        
& 30.81 (4)                        
& 0.22 (5)                        
& 0.21 (5)                        
& 0.21 (5)                        
& 0.23 (4)                        
\\\hline
\raisebox{-5pt}[0pt][0pt]{$9^s$} & \raisebox{-5pt}[0pt][0pt]{Su-Doku} &
\raisebox{-5pt}[0pt][0pt]{$3$}
& 123.40                   
& 127.80                   
& 164.60                   
& 111.93                   
& 108.67                   
& 119.87                   
& 123.93                   
\\\cline{4-10}
 & &.
& 18.89                        
& 19.85                        
& 19.75                        
& 19.10                        
& 19.39                        
& 19.77                        
& 19.96                        
\\\hline
\raisebox{-5pt}[0pt][0pt]{$10^s$} &
\raisebox{-5pt}[0pt][0pt]{TowersOfHanoi}  
&
\raisebox{-5pt}[0pt][0pt]{$5$}
& 145064.20                
& 124222.96                
& 71220.52                 
& 140386.64                
& 97411.80                 
& 134192.96                
& 133760.48                
\\\cline{4-10}
 & &
& 62.43                        
& 46.69                        
& 21.86                        
& 52.19                        
& 32.76                        
& 47.63                        
& 37.60                        
\\\hline
\raisebox{-5pt}[0pt][0pt]{$11^r$} &
\raisebox{-5pt}[0pt][0pt]{TravelingSalesperson}  
&
\raisebox{-5pt}[0pt][0pt]{$5$}
& 2512.20                  
& 1018.80                  
& 3243.16                  
& 2535.40                  
& 1334.32                  
& 2500.16                  
&  947.56                  
\\\cline{4-10}
 & &
& 34.06                        
& 21.63                        
& 42.22                        
& 36.77                        
& 25.70                        
& 34.42                        
& 20.89                        
\\\hline\hline
\multicolumn{3}{|c||}{Average Number of Conflicts}
& 56824.37                 
& 53545.45                 
& 48477.39                 
& 56737.24                 
& 52748.20                 
& 54339.63                 
& 54217.91                 
\\\hline
\multicolumn{3}{|c||}{Average Time (Sum Timeouts)}
& 31.89 (26)                   
& 24.81 (26)                   
& 19.68 (31)                   
& 23.38 (29)                   
& 25.02 (27)                   
& 21.31 (30)                   
& 14.20 (26)                   
\\\hline
\multicolumn{3}{|c||}{Average Penalized Time}
& 49.25                    
& 46.75                    
& 45.28                    
& 48.05                    
& 47.12                    
& 47.14                    
& 37.27                    
\\

%% file: conclusions.tex
\section{Discussion}

We have proposed a number of heuristics for conflict resolution and
conducted a systematic empirical study in the context of our ASP solver \clasp.
However, it is too early to conclude any dominant approach or to make
general recommendations.
As has also been noted in \cite{mitchell05a},
conflict resolution strategies are almost certainly important but have
received little attention in the literature so far.
In fact, dedicated approaches in the SAT area \cite{ryan04a,mafuma04a}
merely aim at reducing the size of recorded nogoods.
Though this might work reasonably well in practice, it is unsatisfactory when
compared to sophisticated decision heuristics 
\cite{golnov02a,ryan04a,mafuma04a,dehana05a}
resulting from more profound considerations.
We thus believe that heuristics in conflict resolution deserve further attention.
Future lines of research may include developing more sophisticated
scoring mechanisms than the ones proposed here, combining several scoring criterions, or
even determining and possibly recording multiple reasons for a conflict
(corresponding to different conflict graphs).
Any future improvements in these directions may significantly boost the state-of-the-art
in both ASP and SAT solving.
%


%% file: paper.bbl
\begin{thebibliography}{}

\bibitem[\protect\citeauthoryear{Baral, Brewka, \& Schlipf}{2007}]{lpnmr07}
Baral, C.; Brewka, G.; and Schlipf, J., eds.
\newblock 2007.
\newblock {\em Proceedings of the Ninth International Conference on Logic
  Programming and Nonmonotonic Reasoning (LPNMR'07)}.
  Springer-Verlag.

\bibitem[\protect\citeauthoryear{Baral}{2003}]{baral02a}
Baral, C.
\newblock 2003.
\newblock {\em Knowledge Representation, Reasoning and Declarative Problem
  Solving}.
\newblock Cambridge University Press.

\bibitem[\protect\citeauthoryear{Bayardo \& Schrag}{1997}]{baysch97a}
Bayardo, R., and Schrag, R.
\newblock 1997.
\newblock Using {CSP} look-back techniques to solve real-world {SAT} instances.
\newblock In {\em Proceedings of the Fourteenth National Conference on
  Artificial Intelligence (AAAI'97)},  203--208.
\newblock AAAI Press/MIT Press.

\bibitem[\protect\citeauthoryear{Beame, Kautz, \& Sabharwal}{2004}]{bekasa04a}
Beame, P.; Kautz, H.; and Sabharwal, A.
\newblock 2004.
\newblock Towards understanding and harnessing the potential of clause
  learning.
\newblock {\em Journal of Artificial Intelligence Research} 22:319--351.

\bibitem[\protect\citeauthoryear{Clark}{1978}]{clark78}
Clark, K.
\newblock 1978.
\newblock Negation as failure.
\newblock In Gallaire, H., and Minker, J., eds., {\em Logic and Data Bases},  293--322.
\newblock Plenum Press.

\bibitem[\protect\citeauthoryear{Dechter}{2003}]{dechter03}
Dechter, R.
\newblock 2003.
\newblock {\em Constraint Processing}.
\newblock Morgan Kaufmann Publishers.

\bibitem[\protect\citeauthoryear{Dershowitz, Hanna, \& Nadel}{2005}]{dehana05a}
Dershowitz, N.; Hanna, Z.; and Nadel, A.
\newblock 2005.
\newblock A clause-based heuristic for {SAT} solvers.
\newblock In Bacchus, F., and Walsh, T., eds., {\em Proceedings of the Eigth
  International Conference on Theory and Applications of Satisfiability Testing
  (SAT'05)}, 
  46--60.
\newblock Springer-Verlag.

\bibitem[\protect\citeauthoryear{E{\'e}n \& S{\"o}rensson}{2003}]{eensor03a}
E{\'e}n, N., and S{\"o}rensson, N.
\newblock 2003.
\newblock An extensible {SAT}-solver.
\newblock In {\em Proceedings of the Sixth International Conference on Theory
  and Applications of Satisfiability Testing ({SAT}'03)},  502--518.

\bibitem[\protect\citeauthoryear{Erdem \& Lifschitz}{2003}]{erdlif03a}
Erdem, E., and Lifschitz, V.
\newblock 2003.
\newblock Tight logic programs.
\newblock {\em Theory and Practice of Logic Programming} 3(4-5):499--518.

\bibitem[\protect\citeauthoryear{Fages}{1994}]{fages94a}
Fages, F.
\newblock 1994.
\newblock Consistency of {C}lark's completion and the existence of stable
  models.
\newblock {\em Journal of Methods of Logic in Computer Science} 1:51--60.

\bibitem[\protect\citeauthoryear{Gebser \bgroup \em et al.\egroup
  }{2007a}]{gekanesc07b}
Gebser, M.; Kaufmann, B.; Neumann, A.; and Schaub, T.
\newblock 2007a.
\newblock clasp: A conflict-driven answer set solver.
\newblock In Baral et~al. \shortcite{lpnmr07},  260--265.

\bibitem[\protect\citeauthoryear{Gebser \bgroup \em et al.\egroup
  }{2007b}]{gekanesc07c}
Gebser, M.; Kaufmann, B.; Neumann, A.; and Schaub, T.
\newblock 2007b.
\newblock Conflict-driven answer set enumeration.
\newblock In Baral et~al. \shortcite{lpnmr07},  136--148.

\bibitem[\protect\citeauthoryear{Gebser \bgroup \em et al.\egroup
  }{2007c}]{gekanesc07a}
Gebser, M.; Kaufmann, B.; Neumann, A.; and Schaub, T.
\newblock 2007c.
\newblock Conflict-driven answer set solving.
\newblock In Veloso, M., ed., {\em Proceedings of the Twentieth International
  Joint Conference on Artificial Intelligence (IJCAI'07)},  386--392.
\newblock AAAI Press/MIT Press.

\bibitem[\protect\citeauthoryear{Gebser \bgroup \em et al.\egroup
  }{2007d}]{contest07a}
Gebser, M.; Liu, L.; Namasivayam, G.; Neumann, A.; Schaub, T.; and
  Truszczy{\'n}ski, M.
\newblock 2007d.
\newblock The first answer set programming system competition.
\newblock In Baral et~al. \shortcite{lpnmr07},  3--17.

\bibitem[\protect\citeauthoryear{Giunchiglia, Lierler, \&
  Maratea}{2006}]{gilima06a}
Giunchiglia, E.; Lierler, Y.; and Maratea, M.
\newblock 2006.
\newblock Answer set programming based on propositional satisfiability.
\newblock {\em Journal of Automated Reasoning} 36(4):345--377.

\bibitem[\protect\citeauthoryear{Goldberg \& Novikov}{2002}]{golnov02a}
Goldberg, E., and Novikov, Y.
\newblock 2002.
\newblock {BerkMin}: A fast and robust {SAT} solver.
\newblock In {\em Proceedings of the Fifth Conference on Design, Automation and
  Test in Europe (DATE'02)},  142--149.
\newblock IEEE Press.

\bibitem[\protect\citeauthoryear{Lee}{2005}]{lee05a}
Lee, J.
\newblock 2005.
\newblock A model-theoretic counterpart of loop formulas.
\newblock In Kaelbling, L., and Saffiotti, A., eds., {\em Proceedings of the
  Nineteenth International Joint Conference on Artificial Intelligence
  (IJCAI'05)},  503--508.
\newblock Professional Book Center.

\bibitem[\protect\citeauthoryear{Lifschitz \& Razborov}{2006}]{lifraz04a}
Lifschitz, V., and Razborov, A.
\newblock 2006.
\newblock Why are there so many loop formulas?
\newblock {\em ACM Transactions on Computational Logic} 7(2):261--268.

\bibitem[\protect\citeauthoryear{Lin \& Zhao}{2004}]{linzha04a}
Lin, F., and Zhao, Y.
\newblock 2004.
\newblock {ASSAT}: computing answer sets of a logic program by {SAT} solvers.
\newblock {\em Artificial Intelligence} 157(1-2):115--137.

\bibitem[\protect\citeauthoryear{Mahajan, Fu, \& Malik}{2005}]{mafuma04a}
Mahajan, Y.; Fu, Z.; and Malik, S.
\newblock 2005.
\newblock Zchaff2004: An efficient {SAT} solver.
\newblock In Hoos, H., and Mitchell, D., eds., {\em Proceedings of the Seventh
  International Conference on Theory and Applications of Satisfiability Testing
  (SAT'04)}, 
  360--375.
\newblock Springer-Verlag.

\bibitem[\protect\citeauthoryear{Marques-Silva \& Sakallah}{1999}]{marsak99a}
Marques-Silva, J., and Sakallah, K.
\newblock 1999.
\newblock {GRASP}: A search algorithm for propositional satisfiability.
\newblock {\em IEEE Transactions on Computers} 48(5):506--521.

\bibitem[\protect\citeauthoryear{Mitchell}{2005}]{mitchell05a}
Mitchell, D.
\newblock 2005.
\newblock A {SAT} solver primer.
\newblock {\em Bulletin of the European Association for Theoretical Computer
  Science} 85:112--133.

\bibitem[\protect\citeauthoryear{Moskewicz \bgroup \em et al.\egroup
  }{2001}]{momazhzhma01a}
Moskewicz, M.; Madigan, C.; Zhao, Y.; Zhang, L.; and Malik, S.
\newblock 2001.
\newblock Chaff: Engineering an efficient {SAT} solver.
\newblock In {\em Proceedings of the Thirty-eighth Conference on Design
  Automation (DAC'01)},  530--535.
\newblock ACM Press.

\bibitem[\protect\citeauthoryear{Ryan}{2004}]{ryan04a}
Ryan, L.
\newblock 2004.
\newblock Efficient algorithms for clause-learning {SAT} solvers.
\newblock Master's thesis, Simon Fraser University.

\bibitem[\protect\citeauthoryear{{Van Gelder}, Ross, \&
  Schlipf}{1991}]{gerosc91a}
{Van Gelder}, A.; Ross, K.; and Schlipf, J.
\newblock 1991.
\newblock The well-founded semantics for general logic programs.
\newblock {\em Journal of the ACM} 38(3):620--650.

\bibitem[\protect\citeauthoryear{Ward \& Schlipf}{2004}]{warsch04a}
Ward, J., and Schlipf, J.
\newblock 2004.
\newblock Answer set programming with clause learning.
\newblock In Lifschitz, V., and Niemel{\"a}, I., eds., {\em Proceedings of the
  Seventh International Conference on Logic Programming and Nonmonotonic
  Reasoning (LPNMR'04)}, 
  302--313.
\newblock Springer-Verlag.

\bibitem[\protect\citeauthoryear{Zhang \bgroup \em et al.\egroup
  }{2001}]{zamamoma01a}
Zhang, L.; Madigan, C.; Moskewicz, M.; and Malik, S.
\newblock 2001.
\newblock Efficient conflict driven learning in a {B}oolean satisfiability
  solver.
\newblock In {\em Proceedings of the International Conference on Computer-Aided
  Design (ICCAD'01)},  279--285.

\end{thebibliography}
